\pdfoutput=1

\documentclass[11pt]{article}
\usepackage[dvipsnames]{xcolor}
\usepackage[]{acl}

\usepackage{amsmath}
\usepackage{amsfonts}
\usepackage{algorithm}
\usepackage{algpseudocode}
\usepackage{graphicx} 

\usepackage{times}
\usepackage{latexsym}
\usepackage{graphicx}
\usepackage{subfigure}
\usepackage{hyperref}       
\usepackage{tabularx}
\usepackage{booktabs} 

\usepackage[T1]{fontenc}

\usepackage[utf8]{inputenc}

\usepackage{microtype}

\usepackage{inconsolata}

\usepackage{array}
\usepackage{tabularx}
\usepackage{adjustbox}
\usepackage{multirow}
\usepackage{enumitem}
\usepackage{xspace}
\usepackage{tcolorbox}
\usepackage{hyperref}
\usepackage{url}
\usepackage{amsmath,amsthm,amsfonts,amssymb,bm,stmaryrd,bbm}
\usepackage[noorphans,vskip=0.75ex,leftmargin=2ex]{quoting}
\usepackage{footmisc}\interfootnotelinepenalty=10000
\usepackage{colortbl}
\usepackage[noabbrev,capitalize]{cleveref}
\newcommand{\headercolor}{\rowcolor{gray!15}}
\newcommand{\ours}{{\sc{RICP}}}

\usepackage{xcolor}         
\usepackage{graphicx}
\usepackage{subcaption}

\newtcolorbox[list inside=prompt,auto counter,number within=section]{prompt}[1][]{
    colbacktitle=black!60,
    coltitle=white,
    fontupper=\footnotesize,
    boxsep=5pt,
    left=0pt,
    right=0pt,
    top=0pt,
    bottom=0pt,
    boxrule=1pt,
    #1,
}

\title{Retrieved In-Context Principles from Previous Mistakes}

\author{
Hao Sun\textsuperscript{1},
Yong Jiang\textsuperscript{2}\thanks{$\quad$ Corresponding Author.},
Bo Wang\textsuperscript{3}\\
\textbf{Yingyan Hou\textsuperscript{4},
Yan Zhang\textsuperscript{1*},
Pengjun Xie\textsuperscript{2},
Fei Huang\textsuperscript{2}}
\\
\textsuperscript{1}Peking University,
\textsuperscript{3}Beijing Institute of Technology\\
\textsuperscript{2}Alibaba Group,
\textsuperscript{4}Chinese Academy of Sciences\\
\tt{sunhao@stu.pku.edu.cn}\\
}

\begin{document}
\maketitle
\begin{abstract}
In-context learning (ICL) has been instrumental in adapting Large Language Models (LLMs) to downstream tasks using correct input-output examples.
Recent advances have attempted to improve model performance through principles derived from mistakes, yet these approaches suffer from lack of customization and inadequate error coverage.
To address these limitations, we propose \textbf{R}etrieved \textbf{I}n-\textbf{C}ontext \textbf{P}rinciples (\textbf{RICP}), a novel teacher-student framework.
In RICP, the teacher model analyzes mistakes from the student model to generate reasons and insights for preventing similar mistakes.
These mistakes are clustered based on their underlying reasons for developing task-level principles, enhancing the error coverage of principles.
During inference, the most relevant mistakes for each question are retrieved to create question-level principles, improving the customization of the provided guidance.
RICP is orthogonal to existing prompting methods and does not require intervention from the teacher model during inference.
Experimental results across seven reasoning benchmarks reveal that RICP effectively enhances performance when applied to various prompting strategies.
\end{abstract}


\section{Introduction}
In recent years, large language models (LLMs) have achieved superior performance on a wide range of reasoning tasks, which include mathematic reasoning \cite{xia2024evaluating, yamauchi2023lpml, imani2023mathprompter, lewkowycz2022solving}, commonsense reasoning \cite{bian2023chatgpt, krause2023commonsense, zhao2024large}, symbolic reasoning \cite{dave2024investigating, kojima2022large, qian2022limitations}, and so on.
To enhance these capabilities further and to align model processing closer to human-like reasoning, recent research has increasingly focused on in-context learning (ICL) \cite{ye2023compositional, shum2023automatic, zhang2022automatic}, where models generate responses based on a few provided correct examples, effectively adapting to new tasks without extensive training.

While ICL has proven effective by primarily leveraging correct examples for task adaptation, learning from mistakes remains a fundamental aspect of human learning processes \cite{edmondson1996learning, chialvo1999learning, berman2006will}.
Recent studies have shown that utilizing mistakes can also enhance the performance of LLMs.
These methods often start by extracting principles from mistakes identified by the models, and then use these principles to avoid future similar mistakes.
For example, TPD \cite{wang2024tpd} uses a teacher model to generate instructions based on the feedback from the student.
LEAP \cite{zhang2024context} creates both low-level and high-level principles from the students' mistakes.
Self-Rethinking \cite{tong2024can} encourages LLMs to rethink whether they have made similar previous mistakes, while TRAN \cite{tong2024can} maintains a rule collection to avoid previous mistakes.

\begin{figure}[t]
  \centering
  \includegraphics[width=1.0\columnwidth]{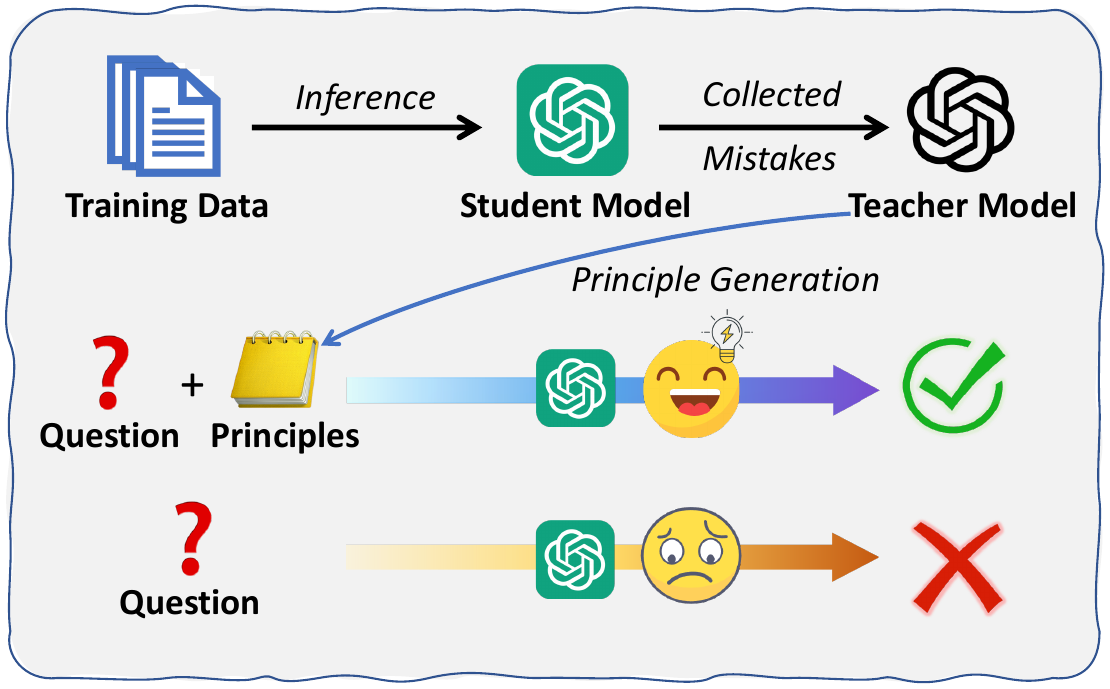}
      \caption{Inference pipeline of \ours{}. The teacher model analyzes mistakes from the student model, creating guiding principles. These principles help prevent the student model from making similar mistakes.}
  \label{fig:introduction}
\end{figure}

However, despite the advancements, these methods still face two main limitations.
(1) Firstly, \textbf{Lack of Customization}.
These methods typically utilize a consistent set of principles across different questions.
However, the fixed principle set may not suit all questions, potentially leading to confusion when inapplicable.
As a result, the performance of LLMs may suffer due to the limited customization of these generalized principles.
(2) Secondly, \textbf{Inadequate Error Coverage}.
Existing methods usually do not adjust the distribution of principles to encompass a broader spectrum of errors.
The inadequate coverage not only narrows the scope of mistakes that the principles can address but also leads to an overemphasis on similar mistakes, resulting in unnecessary token consumption.
Consequently, this limitation not only diminishes the effectiveness of these methods but also increases the inference costs.

To address the aforementioned limitations, we propose a novel approach named \textbf{R}etrieve \textbf{I}n-\textbf{C}ontent \textbf{P}rinciples  (\textbf{RICP}).
\ours{} is a teacher-student framework where the teacher generates principles based on the student’s observed mistakes, and the student applies these principles to prevent the recurrence of similar mistakes.
Specifically, \ours{} involves three stages:
1) \textbf{Insight Generation}: The student model is evaluated on the training set to collect mistakes. The teacher model then analyzes each mistake to generate a high-level reason and several specific insights aimed at avoiding similar errors.
2) \textbf{Principle Formulation}: We cluster mistakes based on their underlying reasons, with each cluster comprising questions that exhibit similar error patterns.
Mistakes sampled from these clusters are analyzed by the teacher model to formulate task-level principles, which are shared by all questions.
For each question, the most similar mistakes are retrieved from each cluster based on question semantics.
We apply clustering to the insights from these mistakes to reduce the redundancy and insights sampled from these clusters establish the question-level principles, which are specific to each question.
3) \textbf{Principle Utilization}: Both the task-level and question-level principles are integrated into the student model’s existing prompt to enhance its question-answering capabilities.

The main advantages of our method are three folds:
(1) Firstly, \ours{} retrieves the most relevant insights for each question, enhancing the specificity and relevance of the question-level principles.
This tailored approach notably \textbf{improves the customization of the principles} provided to the student model, leading to more precise and customized guidance.
(2) Secondly, during the principle formulation stage, we apply reason clustering to construct clusters representing different error patterns, which are then utilized to derive both task-level and question-level principles.
This strategy \textbf{broadens the error coverage of the principles}.
Additionally, insight clustering reduces the redundancy of the question-level principles, enhancing the efficiency of our approach. 
(3) Thirdly, \ours{} is orthogonal to existing prompting methods, enabling seamless integration that improves performance across various prompting strategies. 
Importantly, during inference, \ours{} \textbf{does not require intervention from the teacher model}, which reduces computational overhead.

We have tested \ours{} across seven benchmarks encompassing three reasoning tasks. The empirical results validate that \ours{} can significantly enhance the accuracy of reasoning tasks, confirming its effectiveness and utility in diverse applications.
To summarize, our contributions are as follows:
\begin{itemize}[leftmargin=*]
    \item We propose \ours{}, a novel approach that utilizes teacher-generated principles to prevent the student from making similar mistakes.
    \item \ours{} significantly enhances the customization and error coverage of principles by providing both question-level and task-level principles.
    \item Extensive experiments on seven benchmarks across three reasoning tasks with various LLMs demonstrate that \ours{} consistently enhances model performance.
\end{itemize}

\section{Methodology}
\begin{figure*}[tb]
\includegraphics[width=0.99\textwidth]{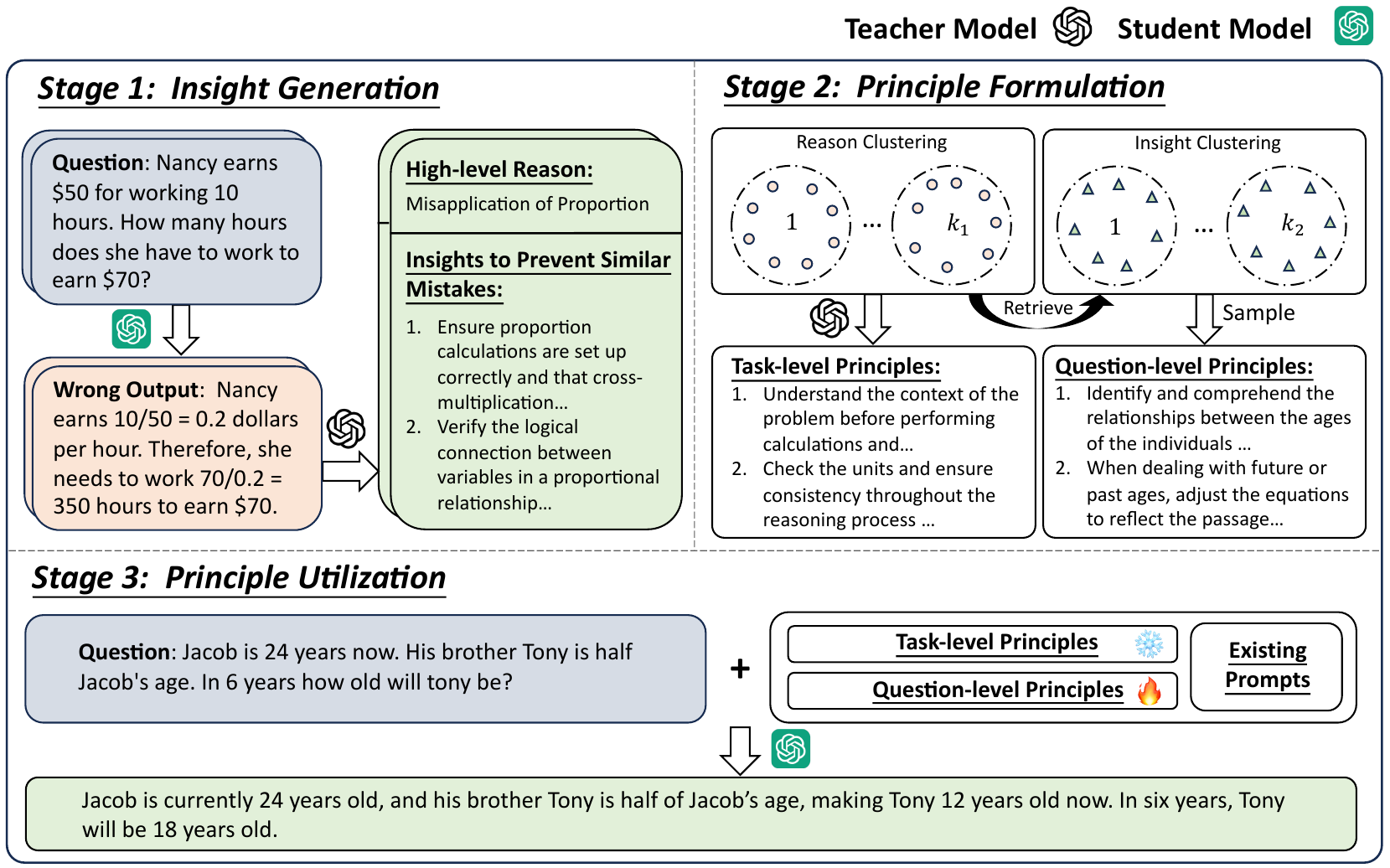}
\centering
\caption{The pipeline of \ours{} includes: 1) Insight Generation: The teacher model analyzes the student model's mistakes and generates high-level reasons and specific insights; 2) Principle Formulation: Task-level principle and question-level principles are generated based on the result of hierarchical clustering;  3) Principle Utilization: principles are integrated into the existing prompt to enhance the performance of the student model.}
\label{fig:model}
\end{figure*}

\subsection{Overview}
In \ours{}, we consider a teacher model and a student model.
The student model first undergoes an examination to collect mistakes, which are then analyzed by the teacher model to obtain high-level reasons and insights to prevent similar mistakes. 
Then, we cluster mistakes based on their underlying reasons, with each cluster comprising questions that exhibit similar error patterns.
Mistakes are sampled from each cluster and analyzed by the teacher model to formulate task-level principles.
Given a question, the most similar mistakes are retrieved from each cluster and the insights of these mistakes are clustered and sampled to formulate question-level principles.
Finally, both task-level and question-level principles are integrated into the student model’s prompt for question answering.

\subsection{Insight Generation}
Given a training dataset $D_{train} = \{ \langle x_i,y_i \rangle\}_{i=1}^{k}$, where $x$ is the question and $y$ is the answer, we ask the student model to undertake an examination on $D_{train}$ to collect its mistakes.
Incorrect solutions are identified by comparing each predicted answer $\hat{y}$ with the ground-truth answer $y$.
\begin{equation}
  D_\text{neg} = \{ (x_i, \hat{r}_i, y_i)  \mid \hat{y}_i \neq y_i, (x_i, y_i) \in D_\text{train} \},
\end{equation}
The collected mistakes reflect the student model's learning status and weakness.

Subsequently, the teacher model analyzes each mistake from $D_\text{neg}$ and generates a high-level reason $R_i$ and a set of insights $I_i$ for each mistake. 
Specifically, for each question $x_i$ in $D_\text{neg}$, we construct the teacher model's prompt incorporating the question $x_i$, the student model’s incorrect rationale $\hat{r}_i$, and the correct answer $y_i$. 
This process yields an insight corpus, defined as:
\begin{equation}
D_\text{insight} = \{ (x_i, R_i, I_i)  \mid  (x_i, \hat{r}_i, y_i) \in D_\text{neg} \},
\end{equation}
Here the reason $R_i$ provides a categorization of the mistake (e.g., “Overlooking Details,” “Misunderstanding of Problem Statement”), while the insights $I_i$ consist of more detailed guidance to help the student model prevent similar mistakes in the future.

\subsection{Principle Formulation}
When formulating the principles, we hope they can be both diverse and customized, which means that the principles can cover as many mistakes as possible but at the same time can provide very precise guidance for each specific question.
To achieve this, we apply hierarchical clustering on the reasons and insights to obtain task-level principles and question-level principles.

\paragraph{Task-level Principles}
To ensure the diversity of principles, we conduct cluster analysis on the mistakes. Specifically, we first compute vector representations for the high-level reasons behind each question in $D_\text{insight}$ using BGE embeddings\footnote{\tiny\url{https://huggingface.co/BAAI/bge-base-en-v1.5}}\cite{bge_embedding}. The reason representations are then processed by the $K$-means clustering algorithm to produce $k$ clusters of questions:
\begin{equation}
  C_i^T = Kmeans(Embed(\{R_j\})),
\end{equation}
where $C_i^T$ represents the $i$-th cluster.
Each cluster consists of questions that share similar error patterns. We then sample questions from each cluster and instruct the teacher to generate task-level principles $ \mathbb{P}_{\textsc{task-level}}$ to prevent similar mistakes.

\paragraph{Question-level Principles}
To ensure the customization of the principles, for a given question, we retrieve the most similar $m$ questions from each cluster $C_i$. Since the insights of these retrieved questions can be highly similar and redundant, we apply $K$-means clustering on the insights:
\begin{equation}
  C_i^Q = Kmeans(Embed(\{I_j\})),
\end{equation}
where $C_i^Q$ represents the $i$-th cluster. 
We then sample $n$ insights from each cluster to serve as the question-level principles $ \mathbb{P}_{\textsc{question-level}}$.

\subsection{Principle Utilization}
During inference, given a question, the task-level and question-level principles are appended to the existing prompt $p$, forming $\mathbb{P}_{\textsc{enhanced}} = \mathbb{P}_{\textsc{task-level}} \oplus \mathbb{P}_{\textsc{question-level}} \oplus p$. The enhanced prompts are then used to answer the question.
The complete algorithm is summarized in \cref{alg}. The generation of task-level principles is a one-time process, while the generation of question-level principles needs to be performed for each question with minimal retrieval and clustering costs. 
Note that our method can be seamlessly integrated into existing prompting methods to achieve further performance improvements without the need for intervention from the teacher model.
\begin{figure}
    \vspace{-3pt}
\centering
\begin{algorithm}[H]
\caption{\ours{} Algorithm}
\label{alg:ricp_alg}
\begin{algorithmic}[1]
\small
\Require Training set $D_{train} = \{ \langle x_i,y_i \rangle\}_{i=1}^{k}$, teacher model $\mathcal{T}$, student model $\mathcal{S}$, Number of Retrieved Questions $m$, Number of Retrieved Insights $n$, Number of Reason Clusters $k_1$, Number of Insight Clusters $k_2$
\For{each QA pair $\langle x_i,y_i \rangle$ in $D_{train}$}
    \State $\hat{r}_i$, $\hat{y}_i \gets \text{FewShotCoT}\left(\mathcal{S}, x_i\right)$
    \State $D_\text{neg} \gets \{ (x_i, \hat{r}_i, y_i)  \mid \hat{y}_i \neq y_i\}$
\EndFor

\For{each $(x_i, \hat{r}_i, y_i)$ in $D_\text{neg}$}
    \State $R_i, I_i \gets \text{InsightGeneration}\left(\mathcal{T}, x_i, \hat{r}_i, y_i\right)$ 
\EndFor
\State $C^T \gets Kmeans(Embed(\{R_i \mid x_i \in D_{train}\}))$
\State $\mathbb{P}_{\textsc{task-level}} \gets \text{PrincipleGeneration}\left(\mathcal{T}, C^T\right)$ 
\State $Q \gets \text{Retrieve}\left(q, C^T\right)$ 
\State $C^Q \gets Kmeans(Embed(\{I_i \mid q_i \in Q\}))$
\State $\mathbb{P}_{\textsc{question-level}} \gets \text{Sample}\left(\mathcal{T}, C^T\right)$ \\
\Return $\mathbb{P}_{\textsc{task-level}}, \mathbb{P}_{\textsc{question-level}}$
\end{algorithmic}
\label{alg}
\end{algorithm}
\end{figure}

\section{Experiment}
\subsection{Experiment Setup}

\paragraph{Baselines}
The baseline prompting methods considered in this work are listed below:
\begin{itemize}[leftmargin=*]
\item \textbf{Standard Prompting} \cite{brown2020language}: The LLM is asked to output the answer directly, without the intermediate reasoning process.
\item \textbf{Chain of Thought} \cite{wei2022chain}: The LLM is instructed to think step-by-step before providing the answer.
\item \textbf{Auto CoT} \cite{zhang2022automatic}: Questions from the training set are first clustered into groups, and the most similar questions are retrieved from each cluster to serve as demonstrations.
\item \textbf{Complex CoT} \cite{fu2022complexity}: Questions from the training set are retrieved based on similarity, and those with the longest rationales are used as demonstrations.
\end{itemize}

\paragraph{Reasoning Tasks}
We conduct experiments on the following reasoning tasks:
    \vspace{-8pt}
\begin{itemize}[leftmargin=*]
    \item \textbf{Mathematical Reasoning}: Four mathematical reasoning datasets are adopted for evaluating: GSM8K \cite{cobbe2021training}, SVAMP \cite{patel2021nlp}, MathQA \cite{amini2019mathqa}, and AQuA \cite{ling2017program}.
    \vspace{-8pt}
    \item \textbf{Commonsense Reasoning}: Two closed-book datasets are employed to evaluate commonsense reasoning: CSQA \cite{talmor2018commonsenseqa} and StrategyQA \cite{geva2021did}.
    \vspace{-8pt}
    \item \textbf{Logical Reasoning}: LogiQA \cite{liu2020logiqa} is used to assess logical reasoning ability.
\end{itemize}
It is worth mentioning that all datasets within the same reasoning task utilize a shared insight corpus.
Please refer to \cref{appendix:dataset}  for more details.

\paragraph{Models} 
We conduct all the experiments using \texttt{GPT-3.5-Turbo} and \texttt{Qwen-Turbo} as student models. For the teacher model, we use \texttt{GPT-4-Turbo}.

\subsection{Main Results}
\begin{table*}[tb]
\centering
\resizebox{\linewidth}{!}{
\begin{tabular}{lcccccccccccc}
\toprule
 & \multicolumn{3}{c}{\textbf{GSM8K}} & \multicolumn{3}{c}{\textbf{SVAMP}} & \multicolumn{3}{c}{\textbf{MathQA}} & \multicolumn{3}{c}{\textbf{AQuA}} \\
\cmidrule (lr){2-4}\cmidrule (lr){5-7}\cmidrule (lr){8-10}\cmidrule (lr){11-13}
\textbf{Method} & \textbf{Vanilla} & \textbf{Ours} & \textbf{Improv} & \textbf{Vanilla} & \textbf{Ours} & \textbf{Improv}  & \textbf{Vanilla} & \textbf{Ours} & \textbf{Improv}  & \textbf{Vanilla} & \textbf{Ours}  & \textbf{Improv} \\
\midrule
\headercolor
\multicolumn{13}{c}{\textbf{Qwen-Turbo}} \\
Standard Prompting & 21.20 & \textbf{23.15} & +1.95 & 61.90 & \textbf{62.60} & +0.70 & 15.30 & \textbf{17.40} & +2.10 & 15.89 & \textbf{17.76} & +1.87  \\
Zero-shot CoT & 75.10 & \textbf{76.10} & +1.00 & \textbf{83.40} & 83.20 & -0.20 &37.60 & \textbf{40.90} & +3.30 & 33.64 & \textbf{38.08} & +4.44  \\
Few-shot CoT & 77.78 & \textbf{79.70} & +1.92 & 82.90 & \textbf{83.50} & +0.60 & 40.60 & \textbf{43.60} & +3.00 & 34.81 & \textbf{38.55} & +3.74  \\
Auto CoT & 76.60 & \textbf{77.80} & +1.20 & 81.70 & \textbf{82.70} & +1.00 & 41.10 & \textbf{42.80} & +1.70 & 36.45 & \textbf{37.15} & +0.70  \\
Complex CoT & 74.60 & \textbf{77.30} & +2.70 & 81.10 & \textbf{82.40} & +1.30 & 40.50 & \textbf{44.00} & +3.50 & 32.24 & \textbf{37.85} & +5.61  \\
\midrule
\headercolor
\multicolumn{13}{c}{\textbf{GPT-3.5-Turbo}} \\
Standard Prompting & 25.60 & \textbf{28.60} & +3.00 & 78.40 & \textbf{79.90} & +1.50 & 21.30 & \textbf{23.20} & +1.90 & 20.33 & \textbf{25.00} & +4.67  \\
Zero-shot CoT &  68.40 & \textbf{74.70} & +6.30 & 68.80 & \textbf{74.40} & +5.60 & 33.50 & \textbf{39.30} & +5.80 & 31.07 & \textbf{38.08} & +7.01  \\
Few-shot CoT &  77.10 & \textbf{78.10} & +1.00 & 80.50 & \textbf{82.70} & +2.20 & 41.30 & \textbf{43.10} & +1.80 & 35.98 & \textbf{40.19} & +4.21  \\
Auto CoT &  74.20 & \textbf{76.70} & +2.50 & 78.40 & \textbf{80.90} & +2.50 & 38.40 & \textbf{40.30} & +1.90 & 36.45 & \textbf{39.02} & +2.57  \\
Complex CoT &  74.20 & \textbf{77.10} & +2.90 & 81.20 & \textbf{83.80} & +2.60 & 38.70 & \textbf{41.60} & +2.90 & 34.35 & \textbf{41.12} & +6.78   \\
\bottomrule
\end{tabular}
}
\caption{Performance comparison of different models on mathematical reasoning benchmarks with and without our enhancement method.}
\label{tab:main_math}
\end{table*}

\begin{table*}[tb]
\centering
\resizebox{\linewidth}{!}{
\begin{tabular}{lcccccccccccc}
\toprule
 & \multicolumn{3}{c}{\textbf{CSQA}} & \multicolumn{3}{c}{\textbf{StrategyQA}} & \multicolumn{3}{c}{\textbf{LogiQA}} & \multicolumn{3}{c}{\textbf{Overall}} \\
\cmidrule (lr){2-4}\cmidrule (lr){5-7}\cmidrule (lr){8-10}\cmidrule (lr){11-13}
\textbf{Method} & \textbf{Vanilla} & \textbf{Ours} & \textbf{Improv} & \textbf{Vanilla} & \textbf{Ours} & \textbf{Improv}  & \textbf{Vanilla} & \textbf{Ours} & \textbf{Improv}  & \textbf{Vanilla} & \textbf{Ours}  & \textbf{Improv} \\
\midrule
\headercolor
\multicolumn{13}{c}{\textbf{Qwen-Turbo}} \\
Standard Prompting & 82.80 & \textbf{83.50} & +0.70 & 66.94 & \textbf{67.55} & +0.61 & 45.45 & \textbf{48.75} & +3.29 & 44.21 & \textbf{45.81} & +1.60  \\
Zero-shot CoT & 87.10 & \textbf{88.10} & +1.00 & 70.92 & \textbf{72.45} & +1.53 & 42.32 & \textbf{47.34} & +5.02 & 61.44 & \textbf{63.74} & +2.30  \\
Few-shot CoT & 82.60 & \textbf{83.70} & +1.10 & 75.31 & \textbf{75.92} & +0.61 & 47.96 & \textbf{49.69} & +1.72 & 63.14 & \textbf{64.95} & +1.81  \\
Auto CoT & \textbf{80.90} & 80.70 & -0.20 & 72.65 & \textbf{73.67} & +1.02 & 49.22 & \textbf{50.94 }& +1.72 & 62.66 & \textbf{63.68} & +1.02   \\
Complex CoT & 81.20 & \textbf{83.40} & +2.20 & 69.18 & \textbf{70.31} & +1.12 & 47.81 & \textbf{50.47} & +2.66 & 60.95 & \textbf{63.68} & +2.73  \\
\midrule
\headercolor
\multicolumn{13}{c}{\textbf{GPT-3.5-Turbo}} \\
Standard Prompting & 72.00 & \textbf{76.00} & +4.00 & 63.78 & \textbf{66.33} & +2.55 & 40.28 & \textbf{43.42} & +3.13 & 45.95 & \textbf{48.92} & +2.97  \\
Zero-shot CoT &  75.60 & \textbf{79.90} & +4.30 & \textbf{67.35} & 67.24 & -0.11 & 42.63 & \textbf{43.73} & +1.10 & 55.34 & \textbf{59.62} & +4.29  \\
Few-shot CoT &  76.90 & \textbf{78.20} & +1.30 & 73.78 & \textbf{76.84} & +3.06 & 42.32 & \textbf{46.08} & +3.76 & 61.13 & \textbf{63.60} & +2.48 \\
Auto CoT &  75.90 & \textbf{78.80} & +2.90 & 71.43 & \textbf{72.04} & +0.61 & 43.57 & \textbf{44.67} & +1.10 & 59.76 & \textbf{61.78} & +2.01 \\
Complex CoT & 76.70 & \textbf{79.30} & +2.60 & 68.57 & \textbf{69.80} & +1.22 & 36.83 & \textbf{38.24} & +1.41 & 58.65 & \textbf{61.57} & +2.92 \\
\bottomrule
\end{tabular}
}
\caption{Performance comparison of different models on commonsense reasoning and logical reasoning benchmarks with and without our enhancement method.}
\label{tab:main_csqa}
\vspace{-8pt}
\end{table*}
In this section, we present a comparison of the performance of existing prompting methods with and without \ours{} in \Cref{tab:main_math} and \Cref{tab:main_csqa}. Based on these results, we analyze as follows:

First, \ours{} consistently improves the performance of all prompting methods across various benchmarks using different LLMs. Notably, when using GPT-3.5-Turbo, \ours{} achieves 22.6\% relative improvement over Zeroshot CoT and 20.0\% relative improvement over Complex CoT on the AQuA dataset. By incorporating both diverse task-level principles and customized question-level principles, our method can provide both comprehensive and precise guidance, thereby significantly improving performance.

Second, the relative improvements among the prompting methods vary. For example, when using GPT-3.5-Turbo, the improvement over zero-shot CoT is 7.5\%, while the improvement over Auto CoT is 3.4\%. This variation is mainly due to the fact that some prompting methods can avoid previous mistakes by learning from demonstrations in the prompt. Consequently, our method offers more significant improvements to methods without demonstrations, eliminating the need for selecting demonstrations. This is further confirmed by the higher performance improvement observed in zero-shot prompting methods like standard prompting and zero-shot CoT.
\begin{figure}[tb]
\includegraphics[width=0.95\linewidth]{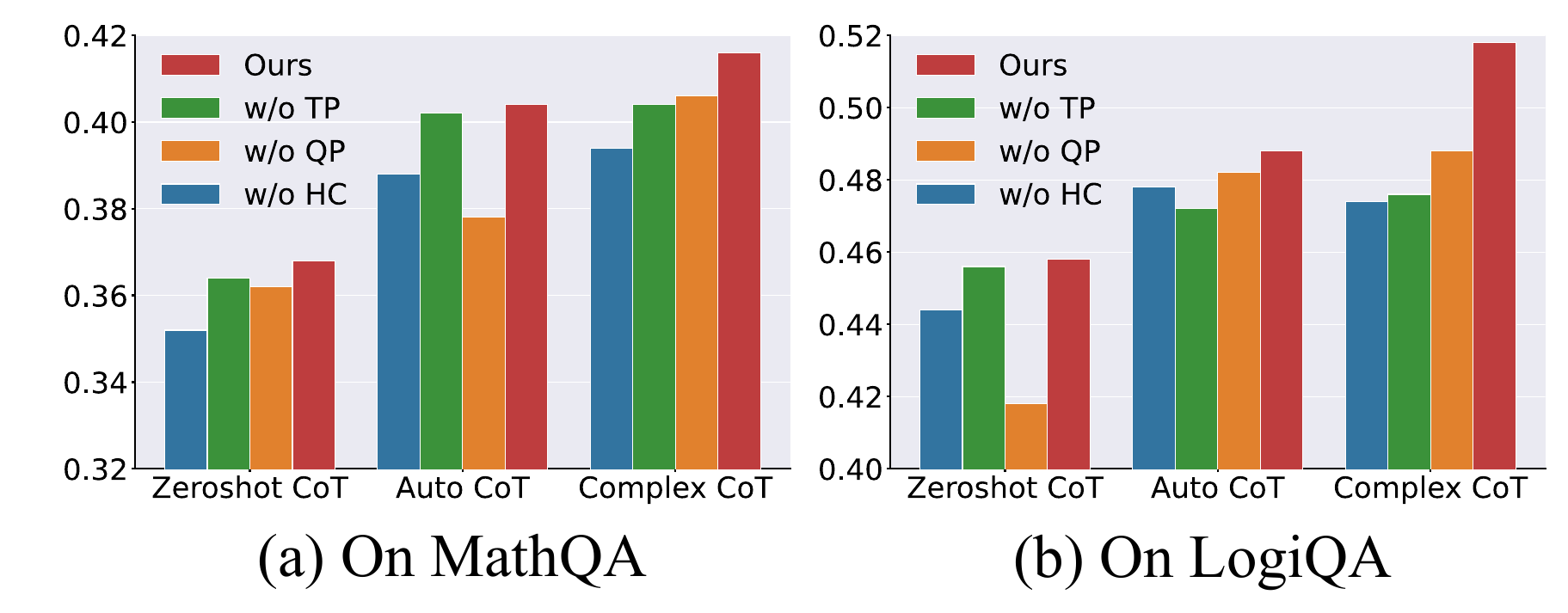}
\centering
\caption{Ablation Study. }
\label{fig:ablation}
\end{figure}

Third, when comparing different LLMs, we find that the relative improvement of our method using GPT-3.5-Turbo is larger than the relative improvement using Qwen-Turbo. This is mainly because the semantic understanding ability of Qwen-Turbo is not as advanced as GPT-3.5-Turbo \cite{bai2023qwen}. As a result, Qwen-Turbo struggles to comprehend and fully utilize the provided principles.

\subsection{Ablation Study}
In this section, we assess the impact of each component in our model on the MathQA and LogiQA datasets using GPT-3.5-Turbo as the LLM.

Results in \Cref{fig:ablation} show that removing any component leads to performance degradation, highlighting their importance.
Notably, the absence of question-level principles leads to a significant decline in performance when applying Auto CoT to the MathQA dataset and Zero-Shot CoT to the LogiQA dataset. 
This is mainly because the question-level principles provide customized guidance for each question, helping the student model avoid previous mistakes effectively.
Conversely, the elimination of task-level principles results in a more modest decrease in performance. This is because, in some cases, the insights provided by task-level principles can be partially compensated by question-level principles. 
However, completely removing task-level principles introduces randomness into the response process when the retrieved question-level principles fail to provide precise guidance for the question.
Moreover, removing hierarchical clustering also leads to a performance drop due to reduced error coverage of the principles. 
This clustering technique is vital for ensuring that the insights are comprehensive, which is crucial for maintaining the model's robust performance.

\subsection{Hyper-parameter Study}
\begin{figure}[tb]
\includegraphics[width=0.98\linewidth]{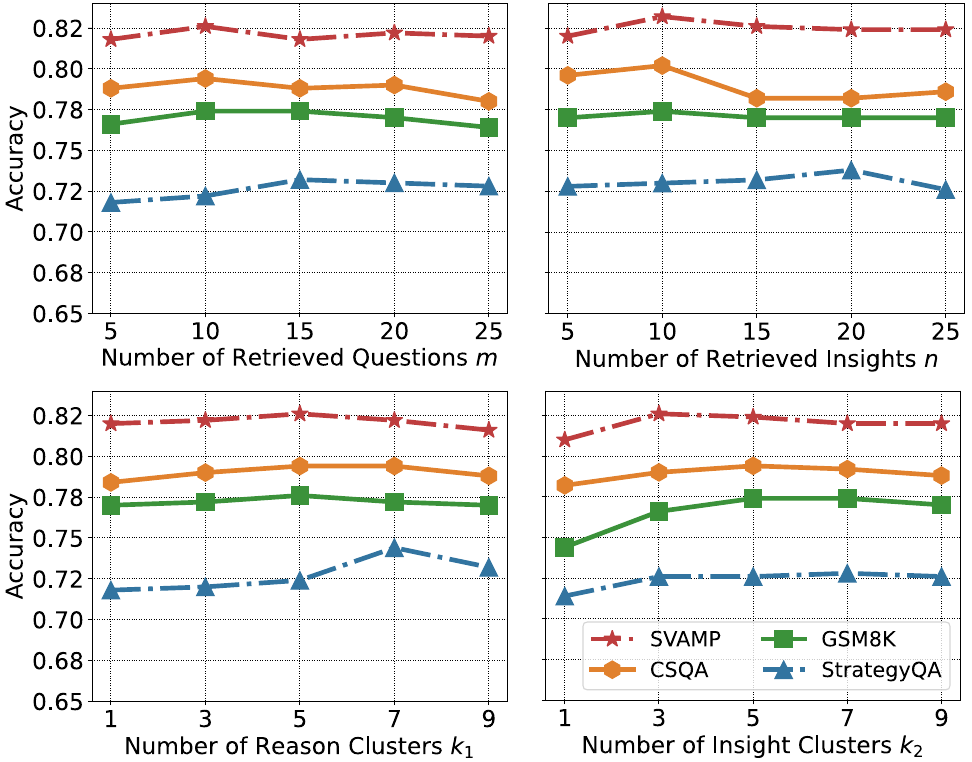}
\centering
\caption{Hyper-parameters}
\label{fig:hyper}
\vspace{-8pt}
\end{figure}
In this section, we investigate hyper-parameters' impact on performance using GSM8K, SVAMP, CSQA and StrategyQA datasets, utilizing Qwen-Turbo as the LLM and Auto CoT as the prompting method. The results are shown in \Cref{fig:hyper}.

We observe a consistent trend across all hyper-parameters: both excessively high and low values adversely affect performance.
Specifically, when the number of retrieved questions is low, the insights provided may not fully meet the current question's requirements. Conversely, a high number introduces noise, which can confuse the model.
With regard to the number of insights, having too few may not sufficiently address the question's needs, while having too many can overwhelm the student model and distract from the most critical insights.
Regarding the number of clusters, too few clusters fail to ensure diversity, while too many clusters can introduce irrelevant information.

\begin{table}[tbp]
\centering
\resizebox{\linewidth}{!}{
\begin{tabular}{lcccccc}
\toprule
\textbf{Model} & \textbf{500} & \textbf{1000} & \textbf{1500} & \textbf{2000} &  \textbf{2500} & \textbf{3000} \\
\midrule
Zero-shot CoT & 44.8 & 45.4 & \textbf{47.4} & 43.8 & 45.2 & 46.2 \\
Few-shot CoT & 48.2 & 48.6 & \textbf{48.7} & 45.8 & 48.0 & 48.2 \\
Auto CoT & 47.0 & 47.8 & \textbf{48.8} & 46.6 & 46.8 & 47.0 \\
Complex CoT & 47.0 & \textbf{50.6} & 48.0 & 49.0 & 48.0 & 49.0\\
\bottomrule
\end{tabular}
}
\caption{Performance comparison of different models on various insight corpus sizes.}
\label{tab:pool_size}
\end{table}

\subsection{Analysis}
\paragraph{Effect of the Size of Insight Corpus}
During the insight generation stage, we collect the student's mistakes and generate corresponding insights. In this section, we analyze the impact of the size of the insight corpus on model performance, using the LogiQA dataset and Qwen-Turbo as the LLM.

From the results shown in \Cref{tab:pool_size}, it is evident that both excessively large and small corpora result in performance decline.
A small corpus may only cover a narrow spectrum of mistakes, limiting the ability to offer guidance across a wide range of questions.
Conversely, an overly large corpus tends to include many similar and redundant insights.
This redundancy provides little new information and can distract the student model from focusing on the most critical insights, thereby leading to a decrease in performance.

\paragraph{Customized Retrieval Analysis}
In this section, we compare the performance of customized retrieval versus random selection on GSM8K and AQuA datasets.
For the random selection method, questions are arbitrarily chosen from the insight pool, and the associated insights are utilized as question-level principles.

\begin{figure}[tb]
\includegraphics[width=0.99\linewidth]{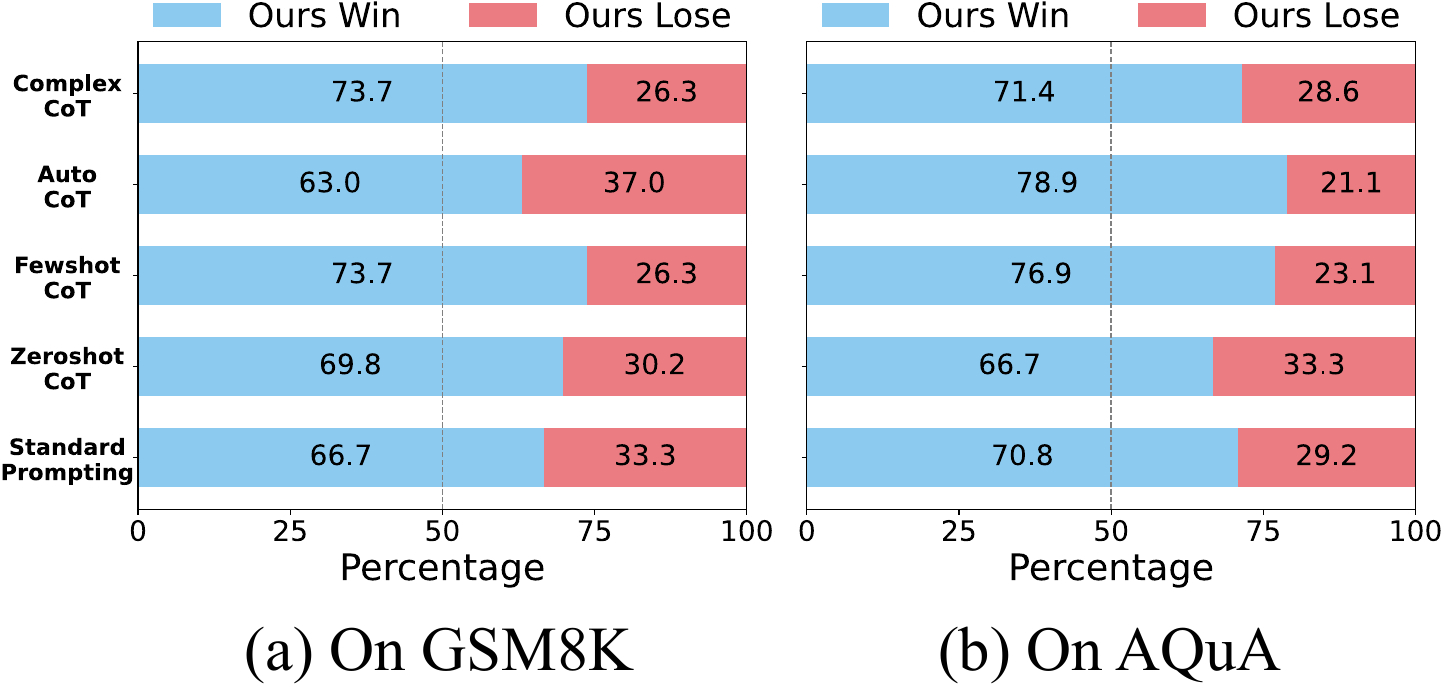}
\centering
\caption{The Comparison between Customized Retrieval and Random Selection.}
\label{fig:win_lose}
\end{figure}

As shown in \Cref{fig:win_lose}, customized retrieval consistently outperforms random selection across all prompting-based methods.
The effectiveness of customized retrieval lies in its ability to provide the most pertinent and suitable guidance for each question, significantly aiding the student model in avoiding previous mistakes.
In contrast, utilizing insights from random selection can sometimes result in performance worse than the vanilla prompting method.
This is mainly because randomly sampled insights may offer guidance that is not applicable to the question. 
Forcing the student model to utilize these inappropriate insights can cause confusion and lead to performance degradation, which further underscores the value of customized retrieval in providing the most pertinent guidance.

\begin{figure}[tb]
\includegraphics[width=0.95\linewidth]{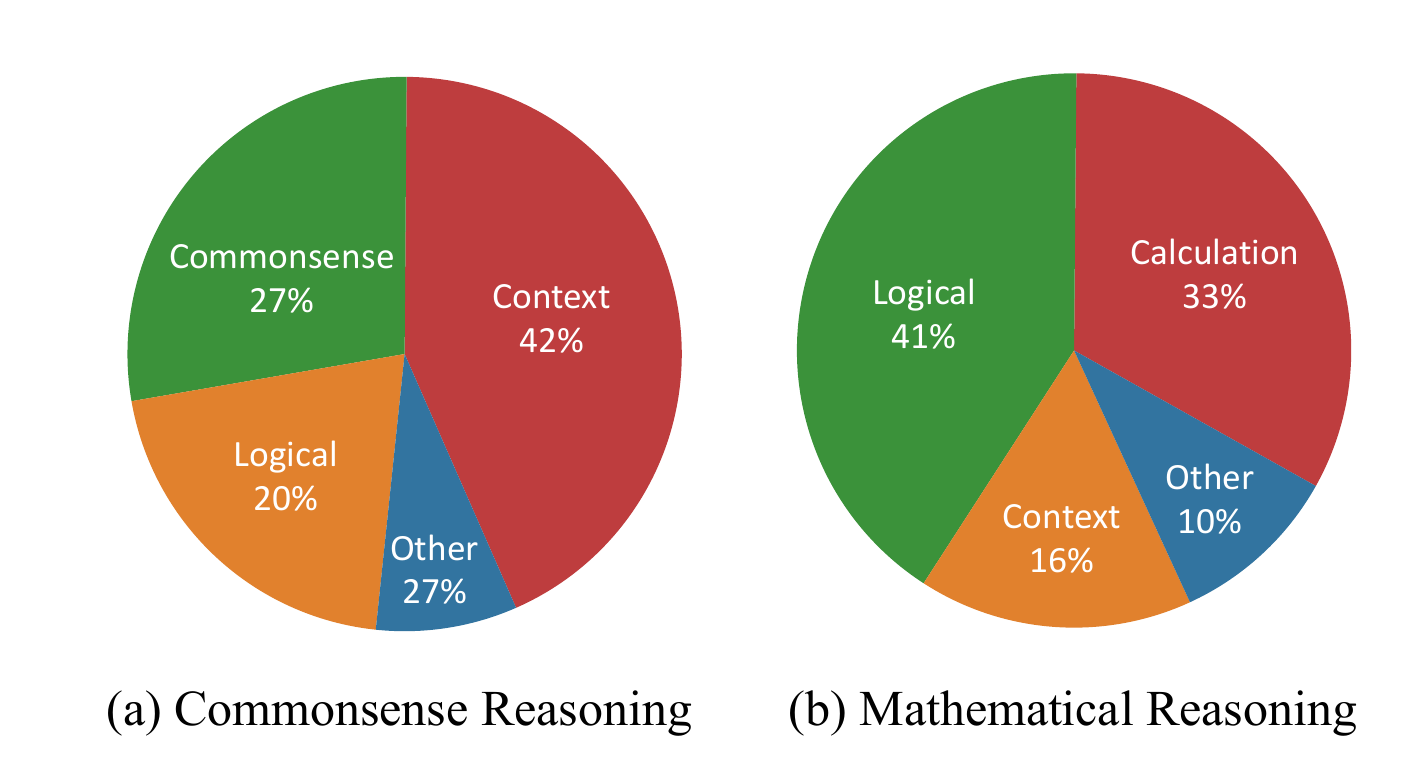}
\centering
\caption{Error Type Distribution}
\label{fig:type_distribution}
\vspace{-8pt}
\end{figure}

\paragraph{Error Type Analysis}
In this section, we investigate the detailed error types of the student model by analyzing mistakes from GSM8K and CSQA datasets, focusing on mathematical and commonsense reasoning. The results are shown in \Cref{fig:type_distribution}.

For commonsense reasoning, the primary errors identified are context and commonsense errors. Context errors often arise from a lack of factual information. Retrieval-augmented Generation (RAG) offers an effective solution by retrieving the most relevant factual information from an external corpus, ensuring the LLM accesses the most up-to-date and accurate data \cite{shi2023replug, guu2020retrieval, jiang2023active}.
For mathematical reasoning, the errors are mainly logical and calculation errors.
Among them, calculation errors can be effectively addressed by providing LLMs with external tools such as calculators \cite{schick2024toolformer, qiao2024autoact, yin2023lumos}.

Despite the absence of an external corpus or tools, our method can still enhance performance by helping the student model avoid previous mistakes. Our approach is particularly effective in addressing logical errors, which are more prevalent in mathematical reasoning than in commonsense reasoning.
This effectiveness is evidenced by the greater relative improvements our method achieves on mathematical reasoning datasets compared to commonsense reasoning datasets.
Additionally, it is worth noting that our method can be seamlessly integrated with RAG and external tools to further enhance model performance.

\subsection{Case Study}
\begin{table*}[!htb]
\centering \footnotesize
\begin{tabular}{p{50em}}
\toprule
\textbf{Case 1: Mathematical Reasoning}
\\ \hline
\vspace{-4pt}
{\color{MidnightBlue}{Question}:} Ashley's pizza delivery costs \$15. What is the total amount that Ashley should give the delivery man if she wants to give a tip that is equal to 1/5 of the amount she ordered?

{\color{MidnightBlue}{CoT Response}:} First, calculate 1/5 of \$15. Thus the answer is that Ashley should give the delivery man \$3 as a tip. \textcolor{cyan}{[WRONG]} 

{\color{MidnightBlue}{Task-level Principle}: }

Principle 1: \textcolor{purple}{Always double-check arithmetic operations.} Simple mistakes in addition, subtraction, multiplication...\\
Principle 3: \textcolor{purple}{Understand the relationships between quantities.} When a problem involves proportional or comparative... 

{\color{MidnightBlue}{Question-level Principle}:}

1.  \textcolor{purple}{Carefully distinguish between costs and revenues to accurately calculate net payment or profit;} take into account...\\
2.  \textcolor{purple}{Verify the inclusion of all components of an equation or formula}, such as base fees or penalties, to ensure...

{\color{MidnightBlue}{CoT+\ours{} Response}:}
To solve this problem, calculate the tip amount, which is 1/5 of the amount she ordered, \$15, resulting in a \$3 tip. Then, add the tip amount to the cost of the pizza delivery. The total amount Ashley should pay is \$15 plus \$3, equaling \$18. Thus, the answer is \$18.\textcolor{cyan}{[CORRECT]}\\

\midrule
\textbf{Case 2: Logical Reasoning} 
\\ \hline
\vspace{-4pt}
{\color{MidnightBlue}{Question}:} 
The raw material of cement is very cheap, like limestone and mud everywhere can be used as raw material of cement. However, the price of cement will be influenced by the price of oil, because it takes a lot of energy to convert raw materials from Pakistan to cement in high-temperature furnaces. what conclusions can be drawn? 

{\color{MidnightBlue}{CoT Response}:}
The conclusion drawn from the given information is that the price of cement is influenced by the price of oil due to the energy required in the conversion process. Therefore,  we can reach a conclusion that the price of cement falls as the price of oil rises. \textcolor{cyan}{[WRONG]} 

{\color{MidnightBlue}{Task-level Principle}:}

Principle 1: \textcolor{purple}{Focus on the main idea and avoid overgeneralization}. When summarizing a text, identify the central theme... \\
Principle 3: \textcolor{purple}{Recognize the implications of specific details.} Pay attention to the details that have significant implications...

{\color{MidnightBlue}{Question-level Principle}:}

1.  \textcolor{purple}{Distinguish between what is explicitly stated and what can be inferred}. Avoid extending conclusions beyond the scope... \\
2.  \textcolor{purple}{Accurately identify and correlate cause-and-effect relationships}; avoid confusing temporal sequence with causality. 

{\color{MidnightBlue}{CoT+\ours{} Response}:} Based on the information that petroleum is the energy source for some high-temperature furnaces for making cement, we can conclude that Petroleum is indeed the energy source for these furnaces. \textcolor{cyan}{[CORRECT]}\\
\bottomrule
\end{tabular}%
\caption{Case studies of solving mathematical and logical reasoning problems with CoT+\ours{} and CoT prompting on the GPT-3.5-Turbo. {\color{MidnightBlue}{Blue}} text indicates the stem, \textcolor{purple}{pink} text indicates the effective hint, \textcolor{cyan}{cyan} text indicates the judgment of whether the answer is correct, \textcolor{cyan}{[CORRECT]} denotes correct, \textcolor{cyan}{[WRONG]} denotes incorrect.}
\label{tab:case_study_part}%
\vspace{-8pt}
\end{table*}%
In this section, we evaluate the effectiveness of our method by analyzing cases from mathematical reasoning and logical reasoning benchmarks.

In the case of mathematical reasoning, the error in the vanilla CoT response stemmed from forgetting to add the original cost of the pizza.
Task-level principles such as ``always double-check arithmetic operations'' and ``understand the proportional relationship'' help prevent such oversights.
Moreover, question-level principles like ``carefully distinguish between costs and revenues'' and ``verify the inclusion of all components'' remind the student model that every element of the calculation should be correctly considered, which is crucial for avoiding the mistake.
In the case of logical reasoning, the error in the vanilla CoT response was due to over-generalization.
Task-level principles such as ``focus on the main idea and avoid over-generalization'' directly point out this issue.
Meanwhile, question-level principles like ``distinguish between what is explicitly stated and what can be inferred'' and ``accurately identify and correlate cause-and-effect relationships'' enhance the accuracy of the logical reasoning process.
These cases demonstrate that by incorporating both task-level and question-level principles, errors in vanilla CoT prompting can be effectively mitigated, thereby verifying the effectiveness of our method.
\section{Related Work}

\subsection{Learning from Mistakes}
Different from machines, humans are able to learn lessons from mistakes, and the lessons can help humans avoid making similar mistakes in the future.
Inspired by this, researchers try to utilize the mistakes to improve the performance of LLMs \cite{li2024turning, an2023learning, wang2024learning}.
\citet{wang2024learning} integrates negative data into the training of language models.
Contrastive Chain-of-Thought \cite{chia2023contrastive} provides both positive and negative examples in the prompt to enhance language model reasoning.
Self-Rethinking \cite{tong2024can} guides LLMs to rethink whether they have made similar previous mistakes.
TRAN \cite{tong2024can} maintains a rule collection to avoid previous mistakes.
The work most related to ours is LEAP \cite{zhang2024context}, which generates low-level and high-level principles from the LLM's mistakes and puts these principles into the prompt.
However, the principles are fixed during inference, which lowers the customization of the guidance, impacting model performance negatively.

\subsection{Teacher-Student Framework}
The teacher-student framework focuses on transferring knowledge from teacher to student.
Traditional finetuning-based methods utilize teacher models to generate training data for student models \cite{rajani2019explain, ho2022large, magister2022teaching}. For instance, \cite{wang2023democratizing} utilizes the teacher model to generate rationales from student's feedback, which are then used for further training of the student. Similarly, \cite{hsieh2023distilling} generates intermediate steps of problem-solving and finetunes the student model on the synthesized data. However, these methods are resource-intensive due to the computational demands of finetuning.
Prompting-based methods leverage teacher models to provide direct guidance to student models \cite{pruthi2022evaluating, saha2023can, yu2023characterizing, wang2024tpd}. For example, \cite{saha2023can} employs the teacher model to deliver customized explanations during testing phases, requiring active teacher involvement.
In contrast, our method can achieve customized guidance without the need for intervention from the teacher model, making it more cost-effective.
\section{Conclusion}
In this paper, we introduce \ours{}, a teacher-student framework designed to prevent the student model from making previous mistakes.
\ours{} significantly enhances the customization and mistake coverage of principles by providing relevant insights for each question and applying hierarchical clustering to the reasons and insights. Extensive experiments across seven benchmarks in three reasoning tasks with various LLMs demonstrate that \ours{} consistently enhances model performance.
\section*{Limitations}
In this paper, we propose a teacher-student framework for preventing the student from making previous mistakes.
The limitations of the proposed method are as follows:

1) Although our model has achieved promising results, it requires the teacher model to be significantly more advanced than the student model to ensure the effectiveness of the generated principles.

2) Despite the fact that our approach does not increase cost and time during inference, there is still some overhead associated with the principle generation process compared to direct few-shot learning, due to the need for pre-constructing the principles.
\section*{Ethics Statement}
This work was conducted in strict compliance with the ACL Ethics Policy. 
All datasets and large language models (LLMs) used for evaluation are publicly available. 
Furthermore, our work aims to explore a reasoning enhancement method. 
We do not foresee any negative ethical impacts arising from our work.
\bibliography{custom}
\clearpage

\appendix
\onecolumn
\section{Dataset Statistics}
\label{appendix:dataset}
\begin{table*}[htbp]
\centering
\resizebox{\linewidth}{!}{
\begin{tabular}{lccccccc}
\toprule
 & \multicolumn{4}{c}{\textbf{Mathematical Reasoning}} & \multicolumn{2}{c}{\textbf{CommonSense Reasoning}} & \multicolumn{1}{c}{\textbf{Logical Reasoning}} \\
 \cmidrule(lr){2-5} \cmidrule(lr){6-7} \cmidrule(lr){8-8}
\textbf{Settings} & \textbf{GSM8K} & \textbf{SVAMP} & \textbf{MathQA} & \textbf{AQuA} & \textbf{CSQA} & \textbf{StrategyQA} & \textbf{LogiQA} \\
 & \cite{cobbe2021training}  & \cite{patel2021nlp} & \cite{amini2019mathqa} & \cite{ling2017program} & \cite{talmor2018commonsenseqa} & \cite{geva2021did} & \cite{liu2020logiqa} \\
\midrule
\headercolor
\multicolumn{8}{c}{\textbf{Dataset statistics}} \\
\#Testing Examples & 1000 & 1000 & 1000 & 428 & 1000 & 980 & 638 \\
\headercolor
\multicolumn{8}{c}{\textbf{Experience Pool}}\\
\#Experience Pool (Qwen-Turbo) & 1118 & 1118 & 1118 & 1118 & 2090 & 2090 & 3437\\
\#Experience Pool (GPT-3.5-Turbo) & 1622 & 1622 & 1622 & 1622 & 2474 & 2474 & 3871\\
\bottomrule
\end{tabular}
}
\caption{Statistics and experimental settings of different tasks/datasets.}
\label{tab:setting}
\end{table*}

All datasets in the same reasoning task share the same insight corpus.

Specifically, in the Mathematical Reasoning task, the training set of GSM8K is utilized to construct the insight corpus, which is subsequently shared by SVAMP, MathQA, and AQuA.

In the Commonsense Reasoning task, we use the training set of CSQA to construct the insight corpus, which is shared by StrategyQA.

In the Logical Reasoning task, since there is only one dataset, we use the training set of LogiQA to construct the insight corpus for itself.

\section{Task-Level Principels}
\begin{figure*}[htbp]
\begin{prompt}[title={Task-level Principles for Mathematical Reasoning using GPT-3.5-Turbo}, label=prompt:search]
Principle 1: Always double-check arithmetic operations. Simple mistakes in addition, subtraction, multiplication, or division can lead to incorrect answers. Ensure that each step of the calculation is performed correctly and consider using a calculator or software to verify results when necessary.
\\\\
Principle 2: Pay attention to units and conversion factors. When dealing with problems that involve different units, such as time, weight, or currency, make sure to convert all quantities to a common unit before performing calculations. This will prevent errors that arise from misinterpreting or mixing units.
\\\\
Principle 3: Understand the relationships between quantities. When a problem involves proportional or comparative relationships, such as "half the price" or "twice as many," ensure that these relationships are applied correctly to the relevant quantities. Misunderstanding these relationships can lead to significant errors in the final answer.
\\\\
Principle 4: Keep track of all elements in the problem. In problems that involve multiple steps or components, it is crucial to account for each part. Missing out on a component or forgetting to include it in the final calculation can result in an incorrect answer.
\\\\
Principle 5: Interpret the context correctly. Ensure that the real-world implications of the problem are understood. This includes recognizing the total quantities involved, the number of entities (people, items, days, etc.), and how these quantities interact with the problem. Misinterpretation of the context can lead to incorrect assumptions and calculations.
\end{prompt}
\end{figure*}

\begin{figure*}[htbp]
\begin{prompt}[title={Task-level Principles for Mathematical Reasoning using Qwen-Turbo}, label=prompt:search]
Principle 1: Understand the context of the problem before performing calculations. Many errors stem from a lack of comprehension of the scenario described in the question. Ensure that the student models grasp the real-world implications of the problem, such as the number of people involved, the time frame, and the quantities being considered.
\\\\
Principle 2: Check the units and ensure consistency throughout the problem. Incorrect solutions often arise from mishandling units, such as days versus weeks or individual items versus groups of items. Student models should be trained to pay close attention to units and convert them appropriately before performing arithmetic operations.
\\\\
Principle 3: Apply the correct arithmetic operations based on the problem's requirements. Misapplication of operations like addition, subtraction, multiplication, and division can lead to incorrect answers. Student models should be guided to identify the correct operation for each step of the problem, considering the relationships between the quantities involved.
\\\\
Principle 4: Double-check the logic of each step in the solution process. Errors can occur when student models make assumptions or skip steps that are crucial for arriving at the correct answer. Encourage the models to review each step for logical consistency and relevance to the problem statement.
\\\\
Principle 5: Practice the distribution and association of mathematical operations. Misunderstandings often occur when dealing with multiple operations and terms. Student models should be adept at applying the distributive property and understanding how to group terms for accurate calculations, especially when dealing with problems that involve proportions or scaling.
\end{prompt}
\end{figure*}

\begin{figure*}[htbp]
\begin{prompt}[title={Task-level Principles for Logical Reasoning using GPT-3.5-Turbo}, label=prompt:search]
Principle 1: Focus on the main idea and avoid overgeneralization. When summarizing a text, identify the central theme without extending the scope to include secondary details or broader concepts that are not the primary focus of the text.
\\\\
Principle 2: Distinguish between descriptive and evaluative statements. Understand when a text is describing a situation, concept, or process versus when it is evaluating or critiquing it. This will help in choosing the most accurate summary or conclusion.
\\\\
Principle 3: Recognize the implications of specific details. Pay attention to the details that have significant implications for the overall argument or narrative of the text. These details often hold the key to understanding the main point or the correct answer.
\\\\
Principle 4: Understand the context of comparative statements. When a text compares two or more items, concepts, or scenarios, ensure that the comparison is correctly interpreted and reflected in the summary or conclusion.
\\\\
Principle 5: Identify the purpose of the text. Determine whether the text aims to inform, persuade, argue, or describe, and use this to guide the selection of the most appropriate summary or answer. This understanding is crucial for accurate comprehension and response.
\end{prompt}
\end{figure*}

\begin{figure*}[htbp]
\begin{prompt}[title={Task-level Principles for Logical Reasoning using Qwen-Turbo}, label=prompt:search]
Principle 1: Distinguish between assumptions and implications. When evaluating statements, it's crucial to differentiate between what is assumed to be true for the scenario to occur and what might be a consequence of the scenario. This helps in identifying the foundational premises required for a situation to exist or an opinion to be valid.
\\\\
Principle 2: Identify the core concept being tested. Focus on the primary subject matter of the question to avoid being misled by peripheral details or related concepts that do not directly answer the question. This ensures that the response is directly relevant to the core concept.
\\\\
Principle 3: Apply logical reasoning to eliminate incorrect options. When faced with multiple choices, use deductive reasoning to rule out options that are inconsistent with the information provided. This process of elimination can often lead to the correct answer by discounting the alternatives that do not fit the given conditions.
\\\\
Principle 4: Understand the definitions of key terms. Ensure that the definitions of critical terms are well understood and applied correctly. Misinterpretation of terms can lead to incorrect conclusions, so it's important to have a clear understanding of the vocabulary used in the question.
\\\\
Principle 5: Analyze the structure of complex problems. Break down complex problems into smaller, more manageable parts to better understand the relationships between different elements. This can help in visualizing the problem and avoiding oversights or misinterpretations that can lead to incorrect answers.
\end{prompt}
\end{figure*}

\begin{figure*}[htbp]
\begin{prompt}[title={Task-level Principles for CommonSense Reasoning using GPT-3.5-Turbo}, label=prompt:search]
Principle 1: Prioritize specificity in problem-solving. When faced with multiple plausible answers, choose the one that is most specific to the question's context. General answers may seem correct but may not be the best fit given the details provided in the question.
\\\\
Principle 2: Consider the practical implications of the question. When determining the correct answer, think about the real-world application and common practices related to the scenario. Avoid answers based solely on assumptions that do not align with typical real-life situations.
\\\\
Principle 3: Avoid overgeneralization. Ensure that the answer selected is not too broad or encompassing when a more precise option is available. Overgeneralization can lead to overlooking the most accurate response.
\\\\
Principle 4: Analyze the question for implicit clues. Some questions contain implicit information that can guide you to the correct answer. Pay attention to subtle hints within the question that may indicate a specific answer over others that are merely correct in a general sense.
\\\\
Principle 5: Distinguish between necessary and sufficient conditions. Understand the difference between what is needed for a situation to occur (necessary) and what alone guarantees the situation (sufficient). Choose the answer that aligns with the necessary condition when the question calls for it.

\end{prompt}
\end{figure*}

\begin{figure*}[htbp]
\begin{prompt}[title={Task-level Principles for CommonSense Reasoning using Qwen-Turbo}, label=prompt:search]
Principle 1: Pay attention to the specificity of language in the problem statement. Avoid overgeneralization and ensure that the answer chosen aligns closely with the specific wording and context provided in the question.
\\\\
Principle 2: Apply logical reasoning to deduce the most appropriate answer. Consider the typical associations and conventional knowledge related to the subject matter to guide the selection of the correct response.
\\\\
Principle 3: Recognize the difference between literal and figurative language. Ensure that the interpretation of the question is based on the literal meaning of the words unless the context clearly indicates a figurative or idiomatic usage.
\\\\
Principle 4: Avoid the influence of personal biases or assumptions that are not supported by the information given in the problem. Base the answer solely on the facts presented in the question, without introducing external assumptions.
\\\\
Principle 5: Understand common knowledge and societal norms as they pertain to the question. Use this understanding to eliminate answers that do not fit within the realm of general consensus or widely accepted facts.
\end{prompt}
\end{figure*}

\section{Case Study}
\begin{table*}[!htb]
\centering \footnotesize
\begin{tabular}{p{50em}}
\toprule
\textbf{Case 1: Mathematical Reasoning} \\
\midrule
\textcolor{blue}{Question}: Ashley's pizza delivery costs \$15. What is the total amount that Ashley should give the delivery man if she wants to give a tip that is equal to 1/5 of the amount she ordered?\\
\textcolor{blue}{CoT Response}: First, calculate 1/5 of \$15, which is \$3. Therefore, the answer is that Ashley should give the delivery man \$3 as a tip. \textcolor{cyan}{[WRONG]} \\
\textcolor{blue}{Task-level Principle}: 

Principle 1: \textcolor{purple}{Always double-check arithmetic operations.} Simple mistakes in addition, subtraction, multiplication, or division can lead to incorrect answers. Ensure that each step of the calculation is performed correctly and consider using a calculator or software to verify results when necessary.

Principle 2: Pay attention to units and conversion factors. When dealing with problems that involve different units, such as time, weight, or currency, make sure to convert all quantities to a common unit before performing calculations. This will prevent errors that arise from misinterpreting or mixing units.

Principle 3: \textcolor{purple}{Understand the relationships between quantities.} When a problem involves proportional or comparative relationships, such as "half the price" or "twice as many," ensure that these relationships are applied correctly to the relevant quantities. Misunderstanding these relationships can lead to significant errors in the final answer.

Principle 4: Keep track of all elements in the problem. In problems that involve multiple steps or components, it is crucial to account for each part. Missing out on a component or forgetting to include it in the final calculation can result in an incorrect answer.

Principle 5: \textcolor{purple}{Interpret the context correctly.} Ensure that the real-world implications of the problem are understood. This includes recognizing the total quantities involved, the number of entities (people, items, days, etc.), and how these quantities interact within the problem. Misinterpretation of the context can lead to incorrect assumptions and calculations.
    
\textcolor{blue}{Question-level Principle}: 

1.  \textcolor{purple}{Carefully distinguish between costs and revenues to accurately calculate net payment or profit;} take into account any necessary deductions or additional charges when determining the final amount.

2.  \textcolor{purple}{Verify the inclusion of all components of an equation or formula}, such as base fees or penalties, to ensure the total calculation reflects the full transaction or scenario.

3.  Pay attention to direct information and avoid unnecessary complexity in calculations. Simplify the problem by establishing basic arithmetic relations from the given data before attempting to solve it.

\textcolor{blue}{CoT+\ours{} Response}:
To solve this problem, calculate the tip amount, which is 1/5 of the amount she ordered, \$15, resulting in a \$3 tip. Then, add the tip amount to the cost of the pizza delivery. The total amount Ashley should pay is \$15 plus \$3, equaling \$18. Thus, the answer is \$18.\textcolor{cyan}{[CORRECT]}\\

\midrule
\textbf{Case 2: Logical Reasoning} \\
\midrule
\textcolor{blue}{Question}: 
The raw material of cement is very cheap, like limestone and mud everywhere can be used as raw material of cement.However, the price of cement will be influenced by the price of oil, because it takes a lot of energy to convert raw materials from Pakistan to cement in high-temperature furnaces.what conclusions can be drawn? \\
\textcolor{blue}{CoT Response}:
The conclusion drawn from the given information is that the price of cement is influenced by the price of oil due to the energy required in the conversion process. Therefore,  we can reach a conclusion that the price of cement falls as the price of oil rises. \textcolor{cyan}{[WRONG]} \\
\textcolor{blue}{Task-level Principle}:

Principle 1: \textcolor{purple}{Focus on the main idea and avoid overgeneralization}. When summarizing a text, identify the central theme without extending the scope to include secondary details or broader concepts that are not the primary focus of the text.

Principle 2: Distinguish between descriptive and evaluative statements. Understand when a text is describing a situation, concept, or process versus when it is evaluating or critiquing it. This will help in choosing the most accurate summary or conclusion.

Principle 3: \textcolor{purple}{Recognize the implications of specific details.} Pay attention to the details that have significant implications for the overall argument or narrative of the text. These details often hold the key to understanding the main point or the correct answer.

Principle 4: Understand the context of comparative statements. When a text compares two or more items, concepts, or scenarios, ensure that the comparison is correctly interpreted and reflected in the summary or conclusion.

Principle 5: Identify the purpose of the text. Determine whether the text aims to inform, persuade, argue, or describe, and use this to guide the selection of the most appropriate summary or answer. This understanding is crucial for accurate comprehension and response.

\textcolor{blue}{Question-level Principle}: 

1.  \textcolor{purple}{Distinguish between what is explicitly stated and what can be inferred}. Avoid extending conclusions beyond the scope of the given information. \\
2.  \textcolor{purple}{Accurately identify and correlate cause-and-effect relationships}; avoid confusing temporal sequence with causality. \\
3.  \textcolor{purple}{Assess all available evidence within the context provided}, resisting the urge to extrapolate beyond the given data. \\

\textcolor{blue}{CoT+\ours{} Response}:
Based on the given information is that petroleum is the energy source of some high-temperature furnaces for making cement, we can conclude that Petroleum is the energy source of some high-temperature furnaces for making cement.
\textcolor{cyan}{[CORRECT]}\\

\bottomrule
\end{tabular}%
\caption{Case studies of solving mathematical reasoning and logical reasoning problems with CoT+RCIP and CoT prompting on the GPT-3.5-Turbo model. \textcolor{blue}{Blue} text indicates the stem, \textcolor{purple}{pink} text indicates the effective hint, \textcolor{cyan}{cyan} text indicates the judgment of whether the answer is correct, \textcolor{cyan}{[CORRECT]} denotes correct, and \textcolor{cyan}{[WRONG]} denotes incorrect.}
\label{tab:case_study_part}%
\end{table*}%

\end{document}